\documentclass[letterpaper]{article} 
\usepackage{aaai2026}  
\usepackage{times}  
\usepackage{helvet}  
\usepackage{courier}  
\usepackage[hyphens]{url}  
\usepackage{graphicx} 
\usepackage{natbib}  
\usepackage{caption} 
\usepackage{algorithm}
\usepackage{algorithmic}
\usepackage{multirow}
\usepackage{xcolor}
\usepackage{colortbl}
\usepackage{amsmath,bm}
\usepackage{amssymb}
\usepackage{pifont}
\usepackage{booktabs}
\newcommand{\xmark}{\ding{55}}%

\usepackage{newfloat}
\usepackage{listings}
\DeclareCaptionStyle{ruled}{labelfont=normalfont,labelsep=colon,strut=off} 
\floatstyle{ruled}
\newfloat{listing}{tb}{lst}{}
\floatname{listing}{Listing}
\title{Stereo 3D Gaussian Splatting SLAM for Outdoor Urban Scenes}
\author{
    Xiaohan Li\textsuperscript{\rm 1}\equalcontrib,
    Ziren Gong\textsuperscript{\rm 2}\equalcontrib,
    Fabio Tosi\textsuperscript{\rm 2},
    Matteo Poggi\textsuperscript{\rm 2},
    Stefano Mattoccia\textsuperscript{\rm 2},
    Dong Liu\textsuperscript{\rm 1},
    Jun Wu\textsuperscript{\rm 3}
}
\affiliations{
    \textsuperscript{\rm 1}University of Science and Technology of China,\\
    \textsuperscript{\rm 2}University of Bologna,\\
    \textsuperscript{\rm 3}Fudan University,

}

\usepackage{bibentry}

\begin{document}

\twocolumn[{
\renewcommand\twocolumn[1][]{#1}
\maketitle
\begin{center} 
    \vspace{-4mm}

    \begin{tabular}{@{\hskip 0pt}c@{\hskip 4pt}c@{\hskip 4pt}c@{\hskip 0pt}} 
        \includegraphics[width=0.3\linewidth]{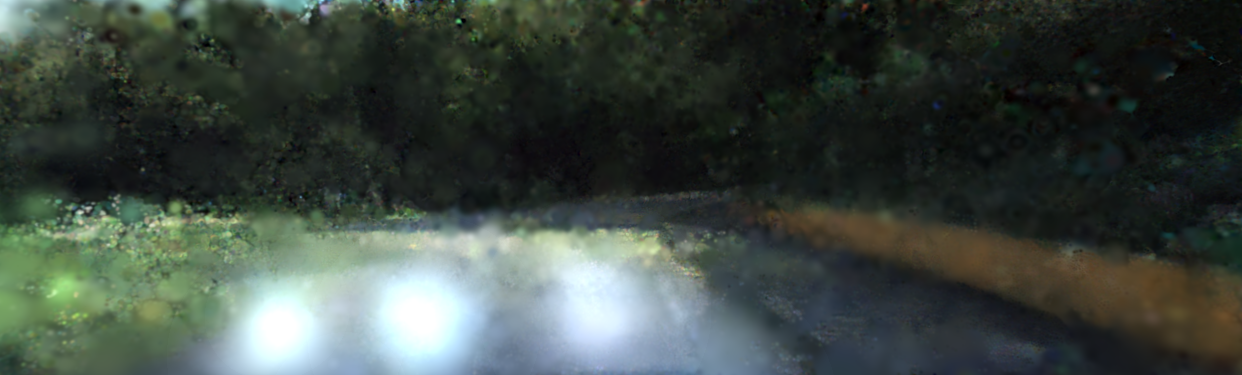} & 
        \includegraphics[width=0.3\linewidth]{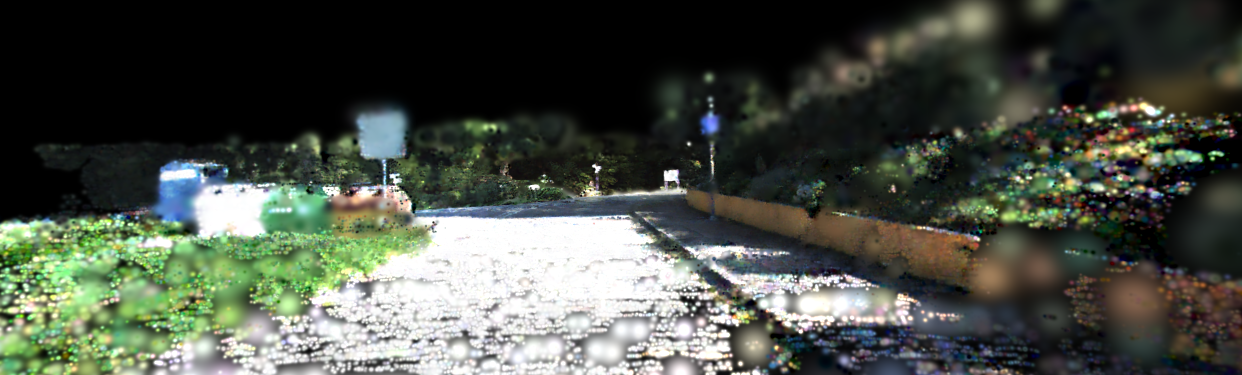} & 
        \includegraphics[width=0.3\linewidth]{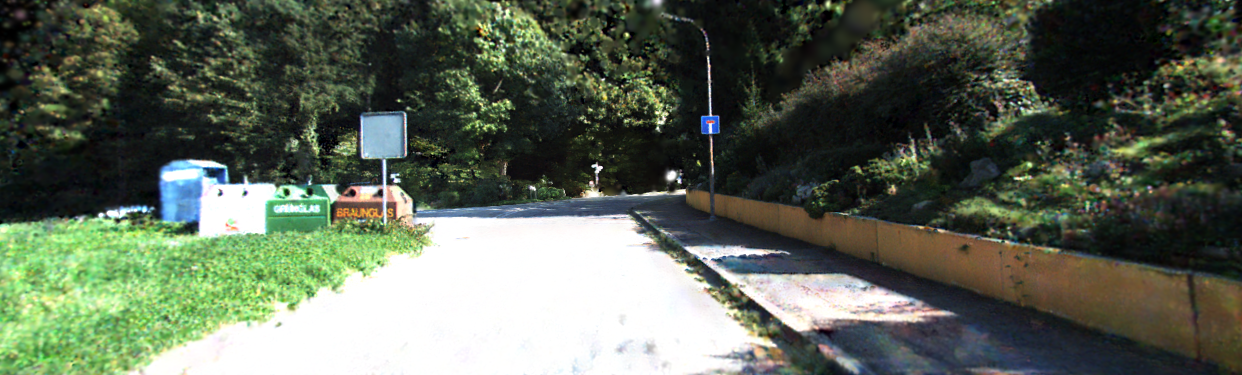} \\

        \includegraphics[width=0.3\linewidth]{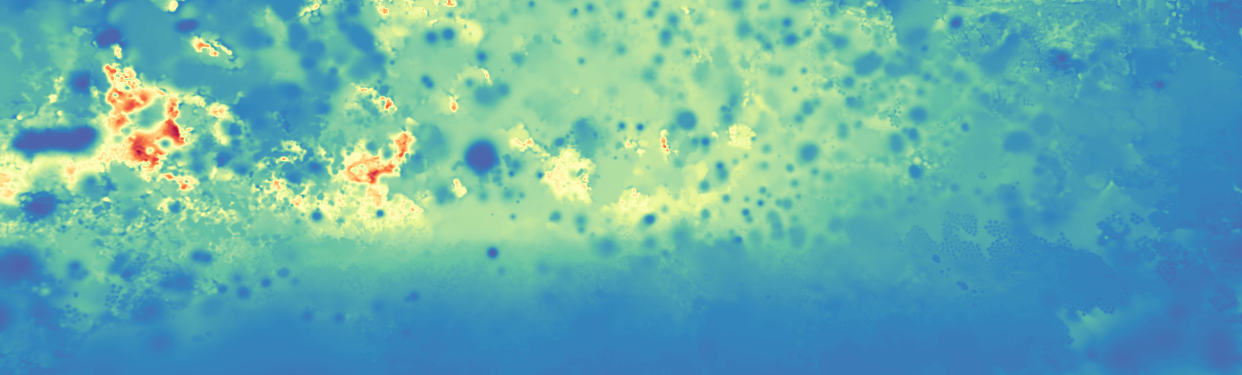} & 
        \includegraphics[width=0.3\linewidth]{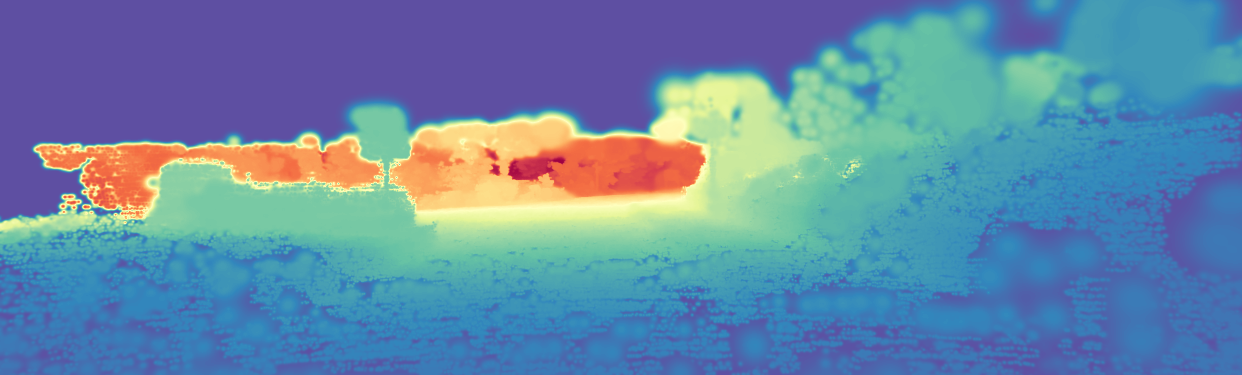} & 
        \includegraphics[width=0.3\linewidth]{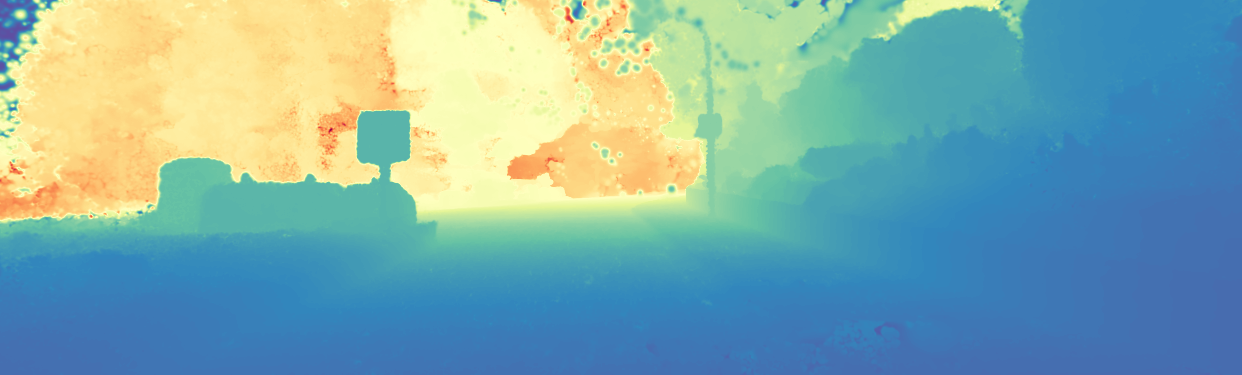} \\ \vspace{-0.1cm} 
        \small SplaTAM \cite{keetha2024splatam} & \small BGS-SLAM (LiDAR) & \small \textbf{BGS-SLAM (Ours)}  \\
    \end{tabular} \vspace{-0.25cm} 
    \captionof{figure}{\textbf{Comparison of Rendering and Depth Estimation.} The top row shows RGB renderings generated by 
    SplaTAM \cite{keetha2024splatam}, our BGS-SLAM method trained with LiDAR depth points, and our approach using only stereo RGB pairs with depth maps from deep stereo networks for supervision. The bottom row presents the corresponding depth renderings. }
    \label{fig:teaser}
\end{center}
}]
\begin{abstract}
3D Gaussian Splatting (3DGS) has recently gained popularity in SLAM applications due to its fast rendering and high-fidelity representation. However, existing 3DGS-SLAM systems have predominantly focused on indoor environments and relied on active depth sensors, leaving a gap for large-scale outdoor applications. We present BGS-SLAM, the first binocular 3D Gaussian Splatting SLAM system designed for outdoor scenarios\footnote{The code will be released in case of acceptance.}. Our approach uses only RGB stereo pairs without requiring LiDAR or active sensors. BGS-SLAM leverages depth estimates from pre-trained deep stereo networks to guide 3D Gaussian optimization with a multi-loss strategy enhancing both geometric consistency and visual quality. Experiments on multiple datasets demonstrate that BGS-SLAM achieves superior tracking accuracy and mapping performance compared to other 3DGS-based solutions in complex outdoor environments.

\end{abstract}

\section{Introduction}
\label{sec:intro}

Simultaneous Localization and Mapping (SLAM), a core research area in computer vision, has been widely applied in autonomous driving, metaverse, and robotics. It primarily utilizes sensor data to estimate the state of a robot while simultaneously constructing an accurate scene representation. Traditional methods \cite{campos2021orb, wang2017stereo, li2024cto} typically formulate this as a maximum a posteriori (MAP) estimation problem, where both robot ego-motion and scene modeling are described as factors in a graph for joint optimization. 

In recent years, neural rendering-based methods have made significant advancements. The emergence of Neural Radiance Fields (NeRF) \cite{mildenhall2021nerf} has profoundly influenced the community by revolutionizing novel view synthesis and scene representation, shifting the focus towards data-driven and differentiable rendering methods. 
Lately, 3D Gaussian Splatting (3DGS) \cite{kerbl20233d} has emerged as a promising alternative. By representing scenes as a collection of 3D Gaussians and leveraging an efficient rasterization strategy, 3DGS achieves fast rendering while providing high-quality scene representation. This naturally aligns with SLAM's requirements for real-time processing and accurate scene reconstruction, making 3DGS-SLAM \cite{keetha2024splatam, matsuki2024gaussian} a rapidly growing research focus in recent years.
However, existing 3DGS-SLAM methods mainly rely on dense and accurate depth maps from RGB-D sensors as training supervision for geometric reconstruction, while also being limited to small-scale indoor scenes. These methods achieve high-fidelity representations in controlled indoor environments but encounter severe challenges in large, complex outdoor settings. 

First, active depth sensors like LiDAR and RGB-D cameras have inherent limitations outdoors. LiDAR systems are expensive, bulky, and power-intensive, making them impractical for many applications. Consumer RGB-D sensors like Microsoft Kinect or Intel RealSense, while more affordable and compact, face even greater limitations outdoors. These devices have significantly shorter effective ranges (typically under 5 meters), and their infrared-based depth sensing becomes unreliable in direct sunlight due to interference with their projected patterns. 
Second, outdoor scenes typically span much larger scales. For example, in the KITTI dataset \cite{geiger2012we}, trajectories often extend over kilometers, causing substantial memory consumption during scene reconstruction and making real-time, efficient large-scale mapping particularly challenging.
Finally, outdoor environments frequently involve drastic viewpoint changes and limited frame overlap, resulting in insufficient optimization constraints. This leads to convergence difficulties and visual artifacts that destabilize training. 

To address these challenges, we propose a novel 3DGS-based architecture specifically designed for large-scale outdoor environments such as autonomous driving scenarios. 
Our approach leverages passive RGB stereo cameras only, which are affordable and lightweight compared to expensive and cumbersome active sensors, and leverages the use of pre-trained deep stereo networks to generate dense depth maps that guide the training of 3D Gaussians, effectively overcoming the lack of reliable depth information in outdoor environments. Additionally, we employ an external tracker based on ORB-SLAM2 \cite{mur2015orb}, which significantly optimizes the entire pipeline and improves the overall system performance.

To the best of our knowledge, we are the first to integrate deep binocular stereo networks with 3DGS-SLAM specifically tailored for outdoor scenarios. 
Compared to the sparse LiDAR point clouds, stereo also provides more complete scene coverage, while demonstrating strong generalization and robustness under challenging lighting \cite{Tosi_IJCV_2025}. Furthermore, conversely to ill-posed, single-view depth estimation approaches \cite{yang2024depthv2, yang2024depth, ke2024repurposing}, stereo still provides proper metric estimates, grounded in epipolar geometry. 
Our experiments confirm that even approximate depth estimations from these networks significantly enhance the optimization process by guiding the positioning of 3D Gaussians and preventing artifacts that typically occur when splats become trapped in incorrect geometric configurations.

In summary, our contributions are the following:  
\begin{itemize}
    \item We propose BGS-SLAM, the first 3D Gaussian Splatting SLAM system for outdoor environments using passive RGB stereo pairs only.
    \item We integrate pre-trained deep stereo networks for dense depth supervision in 3D Gaussian optimization, showing that passive stereo can effectively replace expensive active sensors for outdoor scene reconstruction.
    \item We introduce a combination of normal-based and smoothness losses alongside depth-from-stereo supervision to enhance geometric consistency, reduce artifacts, and improve overall mapping quality.
    \item We present experiments on multiple large-scale outdoor datasets, including KITTI and KITTI-360, demonstrating that our approach significantly surpasses existing 3DGS-SLAM methods in outdoor scenarios, achieving superior tracking, mapping accuracy, and visual quality.
\end{itemize}

\section{Related Work}
\label{sec:formatting}

Our work builds upon neural radiance field-based SLAM \cite{tosi2024nerfs}, particularly focusing on RGB-only methods and 3DGS for outdoor environments.

\textbf{Neural Implicit Representations for SLAM.} Neural implicit representations have revolutionized SLAM research. iMAP \cite{sucar2021imap} pioneered this integration by employing an MLP to map 3D coordinates to color and density. NICE-SLAM \cite{zhu2022nice} addressed scalability through hierarchical representation using multiple pre-trained MLPs. Vox-Fusion \cite{yang2022vox} combined traditional volumetric techniques with neural implicit representations, while Co-SLAM \cite{Wang_2023_CVPR} developed hybrid encodings for robust camera tracking.

For large-scale environments, GO-SLAM \cite{zhang2023go} implemented global optimization techniques including loop closure and bundle adjustment, whereas Point-SLAM \cite{Sandström2023ICCV} introduced a dynamic neural point cloud representation that adapts point density based on scene complexity. Most of these methods, however, rely on RGB-D sensors, limiting outdoor applications.

\textbf{RGB-only SLAM with External Supervision.} RGB-only SLAM methods overcome depth ambiguity through various external supervision signals. DIM-SLAM \cite{li2023dense} employed neural implicit map representation with multi-resolution volume encoding and photometric warping loss. NICER-SLAM \cite{Zhu2023NICER} incorporated monocular depth and normal supervision alongside RGB rendering losses.
iMODE \cite{matsuki2023imode} utilized ORB-SLAM2 for camera pose estimation while enhancing reconstruction through depth-rendered geometry supervision. NeRF-VO \cite{naumann2023nerf} combined DPVO tracking with DPT for depth estimation. Hi-SLAM \cite{zhang2023hi} leveraged DROID-SLAM-based dense correspondence and monocular depth priors to address low-texture environments.
Recent approaches include MoD-SLAM \cite{zhou2024modslam}, which enhanced depth estimation through DPT and ZoeDepth, and MGS-SLAM \cite{zhu2024mgs}, which unified sparse visual odometry with 3DGS through MVS-derived depth supervision. 

\begin{figure*}[!htbp]
    \centering    
    \includegraphics[width=0.95\textwidth]{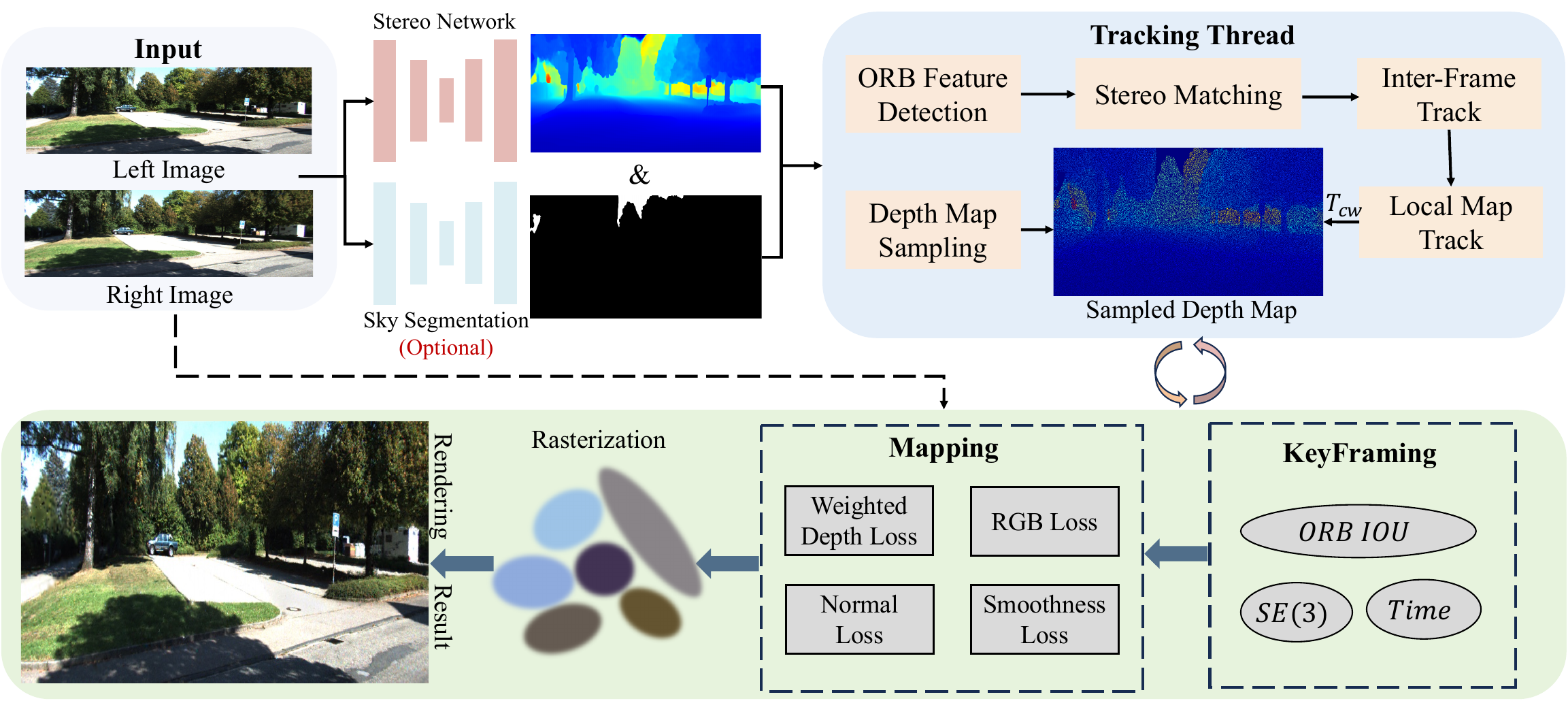}\vspace{-0.25cm}
    \caption{\textbf{Framework Overview.} BGS-SLAM uses stereo images to reconstruct outdoor environments using 3D Gaussians. A pre-trained stereo network extracts dense depth maps from the stereo pairs, with optional sky masking to improve reconstruction. The tracking thread estimates camera poses through feature matching and local bundle adjustment, while the keyframing thread maintains a buffer of key observations. In the mapping thread, a combination of depth, normal, and smoothness losses supervises the 3D Gaussian optimization, enhancing geometric consistency and visual quality of the reconstructed scenes.}
    \label{fig:pipeline}
\end{figure*}

\textbf{3D Gaussian Splatting for SLAM.} 3D Gaussian Splatting \cite{kerbl20233d} offers faster rendering capabilities and improved representation of complex scenes compared to NeRF-based approaches. MonoGS \cite{matsuki2023gaussian} pioneered this paradigm shift by leveraging 3D Gaussians with splatting rendering techniques for a single moving camera. Concurrently, Photo-SLAM \cite{huang2023photo} integrated explicit geometric features with implicit texture representations within a hyper primitives map.
SplaTAM \cite{keetha2024splatam} represented scenes as collections of simplified 3D Gaussians for high-quality color and depth image rendering, while GS-SLAM \cite{yan2023gs} introduced an adaptive expansion strategy and coarse-to-fine tracking technique.
Recent advances include HF-GS SLAM \cite{sun2024high}, which proposed rendering-guided densification strategies, and CG-SLAM \cite{hu2024cg}, which implemented an uncertainty-aware 3D Gaussian field. MM3DGS-SLAM \cite{sun2024mm3dgs} expanded to multi-modal inputs including inertial measurements, while RTG-SLAM \cite{peng2024rtg} addressed large-scale environments by enforcing binary opacity classifications.
For monocular setups, MonoGS++ \cite{monogspp} exploited DPVO \cite{teed2022deep} as an external tracker to estimate initial camera poses. However, in outdoor environments, LIV-GaussMap \cite{hong2024liv} and MM-Gaussian \cite{wu2024mm} still relied on LiDAR sensors for accurate depth measurements. Our work bridges this gap by utilizing stereo RGB images and recent deep stereo matching networks \cite{Tosi_IJCV_2025} to predict depth maps that supervise 3D Gaussian Splatting optimization, enabling high-quality SLAM in outdoor environments without relying on active depth sensors.

\section{Methods}


We detail our binocular BGS-SLAM approach for outdoor environments in this section, with an overview in Fig. \ref{fig:pipeline}.

\subsection{3D Gaussian Splatting (3DGS)}
BGS-SLAM models the scene as a set of 3D Gaussians, denoted as $\mathcal{G} = \{g_1, g_2, \dots, g_N\}$, where $N$ is the number of Gaussians.  Each 3D Gaussian $g_i$ is parameterized by both appearance attributes (color $\mathbf{c}_i$  represented by spherical harmonics and opacity $o_i \in [0,1]$), geometric properties (center position  $\bm{\mu}_i \in \mathbb{R}^3$ and covariance matrix $\mathbf{\Sigma}_i \in \mathbb{R}^{3 \times 3}$) parameters. The spatial influence of each Gaussian is defined as: 
\begin{equation}
    g_i(\mathbf{x}) = e^{-\frac{1}{2} (\mathbf{x}-\bm{\mu}_i)^\top \mathbf{\Sigma}_i^{-1} (\mathbf{x}-\bm{\mu}_i)}
\end{equation}
where the covariance matrix $\mathbf{\Sigma}_i = \mathbf{RSS^{\top}R^{\top}}$ with $\mathbf{S} \in \mathbb{R}^3$ representing the spatial scale and $\mathbf{R} \in \mathbb{R}^{3 \times 3}$ the rotation, parameterized by a quaternion. 

In the rendering process, 3D Gaussians are first projected onto the 2D camera plane as:
\begin{equation}\label{eq2}
    \bm{\mu}' = \mathbf{JW}\bm{\mu}, \quad \mathbf{\Sigma}' = \mathbf{JW \Sigma W^T J^T}
\end{equation}
where $\mathbf{W}$ is the rotational component of the viewing transformation $\mathbf{T}_{cw}$ and $\mathbf{J}$ is the Jacobian matrix which performs linear approximation of the projective transformation. 

With alpha blending, the color and depth at each pixel are generated through: 
\begin{equation}\label{eq3}
  \begin{split}
  C_p = \sum_{i = 1}^{n}\mathbf{c}_{i} \alpha_{i} \prod_{j=1}^{i-1}(1-\alpha_{j}), \quad 
  D_p = \sum_{i = 1}^{n}d_{i} \alpha_{i} \prod_{j=1}^{i-1}(1-\alpha_{j})
  \end{split}
\end{equation}
where the opacity $\alpha_i$ is: 
\begin{equation}
    \alpha_i = o_i \exp\left(-\frac{1}{2} (\mathbf{x}' - \bm{\mu}'_i)^\top \mathbf{\Sigma}'^{-1}_i (\mathbf{x}' - \bm{\mu}'_i)\right)
\end{equation}

During optimization, the parameters of all observed 3D Gaussians are iteratively refined through backpropagation using our mapping losses. For more details, please refer to \cite{kerbl20233d, chen2024survey, tosi2024nerfs}.

\subsection{Deep Stereo Depth Estimation}  

Given a pair of synchronized stereo images, \( I_{\text{left}} \) and \( I_{\text{right}} \), we employ a pre-trained deep stereo network, \( f_{\theta} \), to estimate a dense disparity map \( d \) for outdoor environments:  

\begin{equation}  
d = f_{\theta}(I_{\text{left}}, I_{\text{right}}).  
\end{equation}  

Modern stereo matching networks have evolved into advanced architectures that can be broadly classified into three main categories: Convolutional Neural Network (CNN)-based cost volume aggregation \cite{mayer2016large, kendall2017end}, transformer-based models \cite{li2021revisiting}, and iterative optimization approaches \cite{lipson2021raft}. These networks typically construct a cost volume \( C(d) \) by establishing pixel-wise correspondences between the stereo image pair. Depending on the architecture, this correspondence computation can be performed using correlation layers, absolute or relative feature differences, or direct feature concatenation across the disparity range:  

\begin{equation}  
C(d) = \Psi \big( \Phi(I_{\text{left}}), T_d \big( \Phi(I_{\text{right}}) \big) \big), \quad d \in [d_{\min}, d_{\max}],  
\end{equation}  
where \( \Phi(\cdot) \) denotes a feature extraction function, \( T_d(\cdot) \) represents a disparity-dependent shift operation, and \( \Psi(\cdot) \) defines the cost computation mechanism, which varies based on the network architecture (e.g., feature concatenation, subtraction, or correlation).  

In our framework, we use pre-trained stereo networks that exhibit strong generalization across diverse environments,  e.g., recent foundation models trained on extensive datasets \cite{wen2025foundationstereo, cheng2025monster, bartolomei2025stereo, jiang2025defom}, allowing us to obtain reliable depth estimates without requiring additional domain-specific training. Disparity maps are converted to metric depth \( \tilde{D} \) via the standard stereo triangulation formula:  

\begin{equation}  
\tilde{D} = \frac{f \cdot b}{d},  
\end{equation}  
where \( f \) is the focal length and \( b \) the stereo baseline. These dense depth maps provide rich geometric supervision for our 3DGS optimization process, offering significant advantages over LiDAR-based depth estimation methods due to their higher spatial density and more complete scene coverage.

\subsection{Sky Segmentation}
3DGS often generates ambiguous floaters when rendering sky, leading to a significant number of unnecessary Gaussians and violating multi-view consistency. However, the sky typically occupies only a small portion of outdoor scenes, and the depth map in these regions tends to be inaccurate. 
To address this, we integrate a sky segmentation network \cite{xie2021segformer} into our pipeline, which unifies transformers with a lightweight MLP. Notably, the sky segmentator is optional and designed to enhance system stability.

\subsection{Tracking}

While 3D Gaussian Splatting offers remarkable rendering capabilities, it faces challenges in large-scale outdoor environments due to computational demands and convergence issues. To cope with these issues, we adopt an external tracker based on ORB-SLAM2 \cite{mur2017orb} for several key reasons: (1) feature-based methods like ORB-SLAM2 provide robust real-time tracking even in challenging outdoor conditions with varying illumination and viewpoints, (2) the sparse feature matching approach is computationally efficient compared to the dense optimization required for direct 3D Gaussian optimization, and (3) decoupling the tracking from the mapping thread allows us to maintain stable pose estimation while the 3D Gaussian representation is still being optimized. 

Specifically, at time $i$, the left image $\mathbf{I}_{\mathrm{left}}$ and corresponding right image $\mathbf{I}_{\mathrm{right}}$ are processed into our tracking thread. Following ORB-SLAM2 \cite{mur2017orb}, the ORB features are extracted from both the left and right images which ensures the real-time performance compared to other feature points. Given the camera-inherent $\mathbf{K}$, the ORB features from the left image are searched for matches against those from the right image, yielding a set of stereo correspondences. For the following frames, the incremental camera pose is first initialized under a constant velocity motion model. Then, the matched stereo features are incorporated into a frame-level bundle adjustment (BA). The loss for the frame-level BA can be expressed as:
\begin{equation}\label{eq4}
\min_{\mathbf{T}_i}
\sum_{i \,\in\, \mathcal{O}} \rho\ \Bigl(
\bigl[\mathbf{z}_i - \pi_s(\mathbf{T}_i, \mathbf{X}_i)\bigr]^{T} \,
\Omega_i \,
\bigl[\mathbf{z}_i - \pi_s(\mathbf{T}_i, \mathbf{X}_i)\bigr]
\Bigr),
\end{equation}
where $\mathbf{T}_i$ is the camera pose of current frame $i$, $\mathbf{z}_i$ is the observed stereo measurement, $\mathbf{X}_i$ is the corresponding 3D mappoint, $\pi_s(\cdot)$ is the stereo projection function, $\Omega_i$ is the inverse covariance matrix, and $\rho(\cdot)$ is a robust kernel cost function. Once the incremental ego-motion is estimated, a sliding window of selected keyframes is activated for local bundle adjustment optimization. By constructing a local BA cost function over these keyframes, the tracking thread further reduce the reprojection error and improve the accuracy of ego-motion estimation. Formally, the local BA is:
\begin{gather}
\min_{\{\mathbf{T}_k\},\,\{\mathbf{X}_j\}}
\sum_{k \in \mathcal{K}} \sum_{j \in \mathcal{O}_k}
\rho\Bigl(\mathbf{e}_{kj}^{T}\,\Omega_{kj}\,\mathbf{e}_{kj}\Bigr), \\
\mathbf{e}_{kj} = \mathbf{z}_{kj} - \pi\bigl(\mathbf{T}_k, \mathbf{X}_j\bigr).
\end{gather}
where $\{\mathbf{T}_k\}$ are the keyframe poses within the sliding window, $\{\mathbf{X}_j\}$ are the 3D mappoints visible across keyframes, and $\mathbf{z}_{kj}$ is the observation of point $j$ in keyframe $k$. This local BA jointly optimizes camera poses and 3D structure, providing accurate ego-motion estimates crucial for our mapping thread. Precise pose estimation ensures proper alignment between 3D Gaussians and stereo depth maps, preventing the uncontrolled expansion of Gaussians that would otherwise lead to artifacts and excessive memory consumption in large outdoor scenes.
\subsection{KeyFraming}
In BGS-SLAM, we integrate a keyframing module to ensure the robustness of our system. Most of the recent 3DGS-SLAM systems determine the keyframes according to time intervals. This approach is able to uniformly add keyframes, but the ego-motion in outdoor scenes rarely exhibits linear changes over time and contains intense camera movements. Thus, relying solely on time intervals often leads to inadequate scene overlaps and catastrophic forgetting in outdoor scenes. The keyframing module first assesses the covisibility from the intersection over union (IoU) of the observed ORB keypoints extracted with inter-frames. If the IoU falls below the threshold, the current frame $I_i$ is registered as a new keyframe. To address system instability caused by intense camera movements, the inter-frame transformation is evaluated to determine whether a significant movement occurs. If a large motion change appears, a new keyframe is inserted into the keyframe buffer $\mathcal{T}$. In our paper, the intense camera movement is adaptively defined as 1.5 times the previous motion change:
\begin{equation}\label{eq}
    \mathbf{T_{\mathcal{W} \mathcal{C} }^{t}} \mathbf{(T_{\mathcal{W} \mathcal{C} }^{t-1})^{-1}} > 1.5 \times \mathbf{T_{\mathcal{W} \mathcal{C} }^{t-1}} \mathbf{(T_{\mathcal{W} \mathcal{C} }^{t-2})^{-1}} \rightarrow \mathcal{T} \text{. }
\end{equation}
To maintain computational efficiency, the system retains a limited number of keyframes within the sliding window to optimize both the ego-motion and 3D Gaussian representations. Moreover, a keyframe will be marginalized from the sliding window if the keypoint IoU with the most recent keyframe drops below a specified threshold.

\subsection{Mapping}
The mapping thread receives camera poses from the tracking thread and stereo-estimated depth maps. It represents the scene as a collection of 3D Gaussians optimized through several complementary loss functions:

\textbf{RGB Loss.} We supervise the color reconstruction using a combination of L1 and structural similarity (SSIM) losses:
\begin{align}\label{eq14}
    L_{color} &= \lambda_{rgb} \| I(\mathcal{G}, T_{CW}) - I \|_1 \nonumber \\
              &\quad + \lambda_{ssim} L_{ssim}(I(\mathcal{G}, T_{CW}), I) ,
\end{align}
where $I(\mathcal{G}, T_{CW})$ and $I$ are the rendered and real RGB images, respectively.

\textbf{Weighted Geometric Loss.} While RGB loss is essential, it provides insufficient supervision for outdoor scenes with large textureless regions. Therefore, we introduce a weighted depth loss that incorporates RGB gradient information:
\begin{equation}\label{eq15}
\mathcal{L}_{geo}
\;=\;
g_{\text{rgb}}
\,\frac{1}{\,n\!}\,
\sum \log\Bigl(1 + M\bigl\|D_{s_i} - \hat{D}_{s_i}\bigr\|_1\Bigr)
\end{equation}
where $g_{\text{rgb}} = \exp\bigl(-\nabla I\bigr)$, $\nabla I$ is the RGB image gradient, $n$ is the number of pixels, $M$ is a mask for valid depth values, $D_{s_i}$ is the stereo network-estimated depth map, and $\hat{D}_{s_i}$ is the rendered depth map. 

Based on empirical observations in our experimental validation, we discovered that uniform sampling of stereo depth supervision significantly improves reconstruction quality compared to using the full depth map.  This sampled depth loss is formally defined as:
\begin{equation}\label{eq_sampling}
\mathcal{L}_{\text{geo}}^{\text{sampled}} =  
g_{\text{rgb}} + \frac{1}{|S|}  
\sum_{(i,j) \in S} \log\Bigl(1 + M \bigl| D_{s_{i,j}} - \hat{D}_{s_{i,j}} \bigr|_1 \Bigr)
\end{equation}
where $S \subset \{1,\ldots,H\} \times \{1,\ldots,W\}$ represents a uniform subset of pixel coordinates. For our implementation, we sample approximately $25\%$ of the total pixels using a regular grid pattern. This approach offers several advantages: it reduces the influence of locally correlated errors in stereo depth estimation and promotes smoother optimization by effectively regularizing the supervision signal. 

\textbf{Normal Consistency Loss.} To enhance geometric supervision, we compute normal vectors from both the stereo-estimated and rendered depth maps, and enforce consistency between them: 
\begin{equation}\label{eq16}
\mathcal{L}_{n} \;=\; \frac{1}{\,|{H}|\!}\,\sum \bigl\|{N} - \hat{N}\bigr\|_1,
\end{equation}
where $\hat{N}$ is the rendered normal map, $N$ is the normal map derived from stereo depth estimation, and $H$ denotes the number of patches with valid normal values. 

\textbf{Smoothness Loss.} To ensure geometric consistency, we introduce smoothness loss that penalized abrupt changes in the normal map:
\begin{equation}\label{eq17}
\mathcal{L}_{{s}}
\;=\;
\frac{1}{\,|n|\!}
\sum_{r}\sum_{i,j}
\Bigl(
  \bigl\|\hat{N}_{i+r,j} \;-\; \hat{N}_{i,j}\bigr\|
  \;+\;
  \bigl\|\hat{N}_{i,j+r} \;-\; \hat{N}_{i,j}\bigr\|
\Bigr).
\end{equation}
where $n$ is the number of valid depth values in the rendered depth map.

\textbf{Final Loss.} The final mapping loss combines these components with appropiate weights:
\begin{equation}\label{eq18}
\begin{split}
L_{mapping} &= \lambda_{rgb} L_{rgb} + \lambda_{ssim} L_{ssim} + \lambda_{geo} L_{\text{geo}}^{\text{sampled}} \\  
&\quad + \lambda_{n} L_{n} + \lambda_{s} L_{s}
\end{split}
\end{equation}
where we set $\lambda_{rgb}=0.8$, $\lambda_{ssim}=0.2$, $\lambda_{geo}=0.1$, $\lambda_{n}=0.1$, and $\lambda_{s}=0.5$ in our experiments.

\begin{table}[t]\small
    \centering
    \renewcommand{\arraystretch}{1.2}
    \renewcommand{\tabcolsep}{10pt}
    \resizebox{\linewidth}{!}{%
    \begin{tabular}{ccccc}
    \toprule[2pt]
    Methods & PSNR$\uparrow$ & SSIM$\uparrow$ & LPIPS$\downarrow$ & Depth L1$\downarrow$\\
    \hline
    w/o ss  & 24.19 & 0.92 & 0.11 & 141.71\\
    w/o wd  & 23.86 & 0.91 & 0.12 & 284.36 \\
    w/o nl  & 24.25 & 0.92 & 0.11 & 193.20 \\
    w/o sl  & 23.82 & 0.91 & 0.12 & 186.46\\
    with ds & 21.17 & 0.84 & 0.22 & 263.68 \\
    \textbf{Ours} & \textbf{24.82} & \textbf{0.93} & \textbf{0.10} & \textbf{136.10} \\
    \bottomrule[2pt]
    \end{tabular}
    }\vspace{-0.25cm}
    \caption{\textbf{Ablation Study on the KITTI dataset.} We analyze the effectiveness of sky segmentator (ss), weighted depth loss (wd), normal loss (nl), smoothness loss (sl) and dense depth map supervision (ds) in our proposed SLAM system.}
    \label{Ablation}
\end{table}

\begin{table}[t]\small
    \centering
    \renewcommand{\arraystretch}{1.2}
    \renewcommand{\tabcolsep}{10pt}
    \resizebox{\linewidth}{!}{%
    \begin{tabular}{cccccc}
    \toprule[2pt]
    & Backbones & PSNR $\uparrow$  & SSIM $\uparrow$  & LPIPS $\downarrow$  & Depth L1 $\downarrow$  \\
    \midrule
    & IGEV   & 23.34 & 0.90 & 0.14 & 278.20 \\
    & IGEV++  & 23.19 & 0.89 & 0.15  & 293.57 \\
    & TCSM   & 23.16 & 0.89  & 0.15  & 300.29 \\
    & MonSter-K   & 23.33 & 0.90  & 0.15  & 186.32 \\
    \midrule
    & Mocha   & 24.17 & 0.91  & 0.12  & 462.00  \\
    & FoundationStereo   & 24.57 & 0.92  & 0.11  & \textbf{132.65} \\
    & MonSter-M   & \textbf{24.82} & \textbf{0.93}  & \textbf{0.10}  & 136.10 \\
    \bottomrule[2pt]
    \end{tabular}
    }\vspace{-0.25cm}
    \caption{\textbf{Ablation Study on Stereo Network Selection.} Evaluation of our method's performance using different stereo networks. Depth L1 is in [cm], backbones in the upper part are fine-tuned on KITTI datasets, while backbones in the bottom part are trained on a mix of datasets.} 
    \label{depthNet}
\end{table}

\begin{table*}[!ht]\small
    \centering
    \small
    \scalebox{0.9}{
    \begin{tabular*}{1.0\linewidth}{cccccccccc}
    \toprule[2pt]
    \specialrule{0em}{1pt}{1pt}
    \multirow{2}{*}{\textbf{Methods}} 
      & \multirow{2}{*}{\textbf{Metrics}} 
      & 03 & 05 & 06 & 07 & 09 & 10 
      & \multirow{2}{*}{\textbf{Average}}\\
    \specialrule{0em}{1pt}{1pt}
      & (km/frames)
      & (0.56/801) & (2.2/2761) & (1.2/1101) & (0.69/1101) & (1.7/1591) & (0.92/1201) 
      & \\
    \specialrule{0em}{1pt}{1pt}
    \hline
    \specialrule{0em}{1pt}{1pt}
    {Point-SLAM}\textdagger
    & ATE $\downarrow$ & 81.51 & 104.61 & 170.73 & 79.00 & 138.50 & 102.81 & 112.86  \\
    & PSNR $\uparrow$ & 9.09 & 12.58 & 4.33 & 11.89 & 11.69 & 8.26 & 9.64 \\
    & SSIM $\uparrow$ & 0.30 & 0.48 & 0.24 & 0.47 & 0.37 & 0.38 & 0.37 \\
    & LPIPS $\downarrow$ & 0.74 & 0.66 & 0.89 & 0.64 & 0.71 & 0.69 & 0.72 \\
    & Depth-L1 $\downarrow$ & 227.89 & 428.29 & 405.80 & 211.97 & \textbf{248.12} & 306.25 & 304.72 \\
    \hline
    \specialrule{0em}{1pt}{1pt}
    {SplaTAM}\textdagger
    & ATE $\downarrow$ & 10.20 & 37.13 & 53.78 & 32.82 & 70.23 & 33.96 & 39.69 \\
    & PSNR $\uparrow$ & 14.26 & 14.78 & 16.40 & 16.05 & 15.91 & 14.18 & 15.26 \\
    & SSIM $\uparrow$ & 0.47 & 0.48 & 0.55 & 0.63 & 0.54 & 0.45 & 0.52 \\
    & LPIPS $\downarrow$ & 0.56 & 0.53 & 0.46 & 0.43 & 0.52 & 0.56 & 0.51 \\
    & Depth-L1 $\downarrow$ & 277.91 & 319.99 & 474.61 & 355.67 & 673.33 & 277.86 & 396.89 \\
    \hline
    \specialrule{0em}{1pt}{1pt}
    {MonoGS}\textdagger
    & ATE $\downarrow$ & 57.87 & 51.77 & 92.81 & 51.23 & 81.23 & 61.96 & 66.14 \\
    & PSNR $\uparrow$ & 10.40 & 12.20 & 11.15 & 10.94 & 12.65 & 12.71 & 11.67 \\
    & SSIM $\uparrow$ & 0.25 & 0.37 & 0.28 & 0.38 & 0.42 & 0.38 & 0.35 \\
    & LPIPS $\downarrow$ & 0.71 & 0.65 & 0.76 & 0.67 & 0.71 & 0.68 & 0.70 \\
    & Depth-L1 $\downarrow$ & 681.49 & 403.81 & 575.07 & 568.34 & 666.94 & 674.65 & 595.05 \\
    \hline
    \specialrule{0em}{1pt}{1pt}
    {BGS-SLAM (Lidar)}
    & ATE $\downarrow$  & \textbf{1.77} & \textbf{1.86} & \textbf{0.90} & \textbf{0.60} & \textbf{4.76} & \textbf{3.70} & \textbf{2.26} \\
    & PSNR $\uparrow$ & 11.87 & 6.92 & 10.47 & 8.04 & 8.08 & 10.40 & 9.30 \\
    & SSIM $\uparrow$ & 0.56 & 0.37 & 0.47 & 0.39 & 0.31 & 0.45 & 0.42 \\
    & LPIPS $\downarrow$ & 0.63 & 0.70 & 0.66 & 0.67 & 0.71 & 0.64 & 0.67 \\
    & Depth-L1 $\downarrow$ & 234.77 & 286.10 & 303.22 & 273.60 & 381.24 & 256.28 & 289.20 \\
    \hline
    \specialrule{0em}{1pt}{1pt}
    {\textbf{BGS-SLAM (Ours)}}
    & ATE $\downarrow$ & \textbf{1.77} & \textbf{1.86} & \textbf{0.90} & \textbf{0.60} & \textbf{4.76} & \textbf{3.70} & \textbf{2.26} \\
    & PSNR $\uparrow$ & \textbf{24.82} & \textbf{19.16} & \textbf{23.57} & \textbf{20.14} & \textbf{18.99} & \textbf{22.56} & \textbf{21.54} \\
    & SSIM $\uparrow$ & \textbf{0.93} & \textbf{0.72} & \textbf{0.87} & \textbf{0.78} & \textbf{0.73} & \textbf{0.85} & \textbf{0.81} \\
    & LPIPS $\downarrow$ & \textbf{0.10} & \textbf{0.29} & \textbf{0.15} & \textbf{0.22} & \textbf{0.30} & \textbf{0.18} & \textbf{0.21} \\
    & Depth-L1 $\downarrow$ & \textbf{136.10} & \textbf{253.03} & \textbf{226.89} & \textbf{192.99} & 371.00 & \textbf{161.70} & \textbf{223.62} \\
    \bottomrule[2pt]
    \end{tabular*}}\vspace{-0.25cm}
    \caption{\textbf{Quantitative Evaluation on the KITTI dataset.} Our BGS-SLAM is evaluated on the whole image recorded on the sequences. Methods indicated with \textdagger fail to process the entire recorded image and therefore, their performance is reported on the first 300 frames of all sequences. MonoGS is reported in RGB-D mode. ATE RMSE [m]$\downarrow$, Depth L1 [cm]$\downarrow$ and bold numbers indicate the best result.}\vspace{-0.25cm}
    \label{results}
  \end{table*}

\section{Experiments}

\subsection{Experimental Setup}
\subsubsection{Datasets.} We evaluate BGS-SLAM on the KITTI \cite{Geiger2012CVPR} and the KITTI-360 datasets \cite{liao2022kitti}. Both datasets provide rich sensor data from a vehicle platform with stereo cameras, Velodyne LiDAR, GPS, and IMU, covering diverse driving scenarios including urban areas, residential streets, and highways under varying illumination conditions. 
We focus on the KITTI Odometry split, which contains 22 sequences. Among them, we randomly select 6 sequences with ground truth poses, with trajectory lengths ranging from 0.56 km to 2.2 km. We additionally evaluate on KITTI-360, which offers expanded coverage and complexity, selecting multiple sequences with different scene scales and dynamics to validate BGS-SLAM's robustness and scalability under more complex outdoor conditions.

\subsubsection{Evaluation Metrics.}
\label{sec:metrics}
We evaluate BGS-SLAM on tracking and mapping. For tracking, we report the RMSE of Absolute Trajectory Error (ATE). For rendering quality, we follow radiance-field-based SLAM methods and report PSNR, SSIM \cite{wang2004image}, and LPIPS \cite{zhang2018unreasonable}. Geometric accuracy is measured via Depth L1 error between rendered depth maps and LiDAR ground truth.

\subsubsection{Implementation Details.}
All experiments, including BGS-SLAM and baselines, are run on a desktop with Intel(R) Xeon(R) Gold 6326 CPU @ 2.90GHz and a NVIDIA A40 GPU with 48Gb memory. We adopt  ORB-SLAM2 \cite{mur2017orb} as the external tracker for robust pose estimation. For stereo depth, we use the publicly available MonSter \cite{cheng2025monster} network with pre-trained weights. For learning rate in 3DGS mapping, color is set to \( 2.5 \times 10^{-3} \), rotation and scale to \( 10^{-3} \), opacity to \( 5 \times 10^{-2} \), and the opacity removal threshold to \( 5 \times 10^{-3} \). All results are averaged over three runs.

\begin{table*}[!h]\small
    \centering
    \small
    \scalebox{0.9}{
    \begin{tabular*}{1.0\linewidth}{cccccccccc}
    \toprule[2pt]
    \specialrule{0em}{1pt}{1pt}
    \multirow{2}{*}{\textbf{Methods}} 
      & \multirow{2}{*}{\textbf{Metrics}} 
      & 0002 & 0004 & 0005 & 0007 & 0008 & 0009 
      & \multirow{2}{*}{\textbf{Average}}\\
    \specialrule{0em}{1pt}{1pt}
      & (km/frames)
      & (11.5/14k) & (9.97/11.6k) & (4.69/6.7k) & (4.89/3.4k) & (7.13/8.8k) & (10.58/14k) 
      & \\
    \specialrule{0em}{1pt}{1pt}
    \hline
    \specialrule{0em}{1pt}{1pt}
    {Point-SLAM}\textdagger
    & ATE $\downarrow$ & 99.56 & 161.56 & 56.07 & 247.38 & 99.53 & 159.80 & 137.32  \\
    & PSNR $\uparrow$ & 12.65 & 7.90 & 12.90 & 12.21 & 7.16 & 5.83 & 9.77 \\
    & SSIM $\uparrow$ & 0.47 & 0.38 & 0.43 & 0.41 & 0.18 & 0.26 & 0.35 \\
    & LPIPS $\downarrow$ & 0.69 & 0.70 & 0.67 & 0.73 & 0.92 & 0.89 & 0.77 \\
    & Depth-L1 $\downarrow$ & 371.20 & 485.35 & 720.09 & 746.07 & 494.01 & 676.43 & 582.19 \\
    \hline
    \specialrule{0em}{1pt}{1pt}
    {SplaTAM}\textdagger
    & ATE $\downarrow$ & 56.19 & 67.55 & 23.96 & 138.98 & 58.12 & 57.13 & 66.99 \\
    & PSNR $\uparrow$ & 12.53 & 12.39 & 12.24 & 13.02 & 13.33 & 11.94 & 12.57 \\
    & SSIM $\uparrow$ & 0.29 & 0.34 & 0.31 & 0.34 & 0.36 & 0.40 & 0.34 \\
    & LPIPS $\downarrow$ & 0.60 & 0.57 & 0.61 & 0.58 & 0.58 & 0.58 & 0.59 \\
    & Depth-L1 $\downarrow$ & 492.51 & 586.81 & 684.68 & 727.28 & 501.30 & 724.98 & 619.59 \\
    \hline
    \specialrule{0em}{1pt}{1pt}
    {MonoGS}\textdagger
    & ATE $\downarrow$ & 43.91 & 79.70 & 31.11 & 177.12 & 52.08 & 103.45 & 81.23 \\
    & PSNR $\uparrow$ & 11.23 & 12.10 & 11.00 & 11.43 & 10.63 & 11.12 & 11.25 \\
    & SSIM $\uparrow$ & 0.32 & 0.38 & 0.33 & 0.32 & 0.30 & 0.47 & 0.35 \\
    & LPIPS $\downarrow$ & 0.73 & 0.69 & 0.70 & 0.68 & 0.69 & 0.67 & 0.69 \\
    & Depth-L1 $\downarrow$ & 616.64 & 681.48 & 818.60 & 797.51 & 629.82 & 880.14 & 737.36 \\
    \hline
    \specialrule{0em}{1pt}{1pt}
    {BGS-SLAM (Lidar)}
    & ATE $\downarrow$  & \textbf{3.43} & \textbf{3.25} & \textbf{2.81} & \textbf{2.77} & \textbf{5.55} & \textbf{6.28} & \textbf{4.01} \\
    & PSNR $\uparrow$ & 10.54 & 10.27 & 10.99 & 10.21 & 10.07 & 9.04 & 10.19 \\
    & SSIM $\uparrow$ & 0.40 & 0.40 & 0.38 & 0.39 & 0.36 & 0.32 & 0.38 \\
    & LPIPS $\downarrow$ & 0.68 & 0.64 & 0.68 & 0.66 & 0.69 & 0.71 & 0.68 \\
    & Depth-L1 $\downarrow$ & \textbf{291.01} & 312.82 & 407.84 & \textbf{387.22} & 385.43 & 445.84 & 371.69 \\
    \hline
    \specialrule{0em}{1pt}{1pt}
    {\textbf{BGS-SLAM (Ours)}}
    & ATE $\downarrow$ & \textbf{3.43} & \textbf{3.25} & \textbf{2.81} & \textbf{2.77} & \textbf{5.55} & \textbf{6.28} & \textbf{4.01} \\
    & PSNR $\uparrow$ & \textbf{23.24} & \textbf{24.68} & \textbf{24.93} & \textbf{24.29} & \textbf{24.36} & \textbf{20.47} & \textbf{23.66} \\
    & SSIM $\uparrow$ & \textbf{0.87} & \textbf{0.90} & \textbf{0.91} & \textbf{0.88} & \textbf{0.89} & \textbf{0.80} & \textbf{0.87} \\
    & LPIPS $\downarrow$ & \textbf{0.18} & \textbf{0.14} & \textbf{0.14} & \textbf{0.17} & \textbf{0.15} & \textbf{0.26} & \textbf{0.17} \\
    & Depth-L1 $\downarrow$ & 314.80 & \textbf{215.93} & \textbf{285.88} & 457.50 & \textbf{306.47} & \textbf{428.89} & \textbf{334.91} \\
    \bottomrule[2pt]
    \end{tabular*}}\vspace{-0.25cm}
    \caption{\textbf{Quantitative Evaluation on the KITTI-360 dataset.} Our BGS-SLAM is evaluated on the whole image recorded on the sequences. Methods indicated with \textdagger fail to process the entire image and is reported on the first 300 frames of all sequences. MonoGS is reported in RGB-D mode. Note that in the "(km/frames)" row, "k" is used as a shorthand for 1,000 frames.}\vspace{-0.25cm}
    \label{results-360}
  \end{table*}
  

\subsection{Ablation Study}
\textbf{Component Analysis.} In Table \ref{Ablation}, we present ablation experiments on a KITTI sequence to validate each component of our approach. Our full system achieves the best overall performance across all metrics. Removing the sky segmentation module (``w/o ss") leads to decreased visual quality metrics (PSNR -0.63dB) by introducing inaccurate supervision from sky regions. Without the weighted depth loss (``w/o wd"), depth accuracy deteriorates substantially (Depth L1 increases by 148.26 cm), while maintaining reasonable visual quality, highlighting its importance for geometric reconstruction. The absence of normal loss (``w/o nl") or smoothness loss (``w/o sl") results in increased depth errors and slightly reduced rendering quality, confirming their role in enhancing structural details. Using dense depth maps without our selective supervision strategy (``with ds") performs worse, demonstrating that balancing supervision signals is crucial. 

\textbf{Stereo Network Analysis.} Table~\ref{depthNet} reports the performance of BGS-SLAM integrated with various state-of-the-art stereo matching networks. The upper section of the table includes models fine-tuned on the KITTI dataset (IGEV~\cite{xu2023iterative}, IGEV++\cite{xu2024igev++}, TCSM\cite{zeng2024temporally}, and MonSter-K \cite{cheng2025monster}, while the lower section includes models trained on a broader mix of datasets (Mocha \cite{chen2024mocha}, FoundationStereo \cite{wen2025foundationstereo}, and MonSter-M \cite{cheng2025monster}). Notably, compared with KITTI-only models, networks trained on multiple datasets exhibit overall superior performance across both rendering quality metrics and geometric accuracy. In particular, MonSter-M achieves the best PSNR (24.82), while maintaining a low depth error (136.10 cm), significantly outperforming models such as IGEV (278.20 cm) and TCSM (300.29 cm). Furthermore, these multi-dataset models exhibit stronger zero-shot generalization capability, which is critical for long-term SLAM deployment in unseen environments. We therefore adopt MonSter-M as our default stereo network for optimal accuracy-generalization trade-off.

\begin{figure}[t]
\centering
\setlength{\tabcolsep}{5pt}
\begin{tabular}{@{\hskip 0pt}c@{\hskip 1pt}c@{\hskip 1pt}c@{\hskip 0pt}}

\rotatebox{90}{\hspace{0.2em} \tiny{SplaTAM }} &
\includegraphics[width=0.23\textwidth]{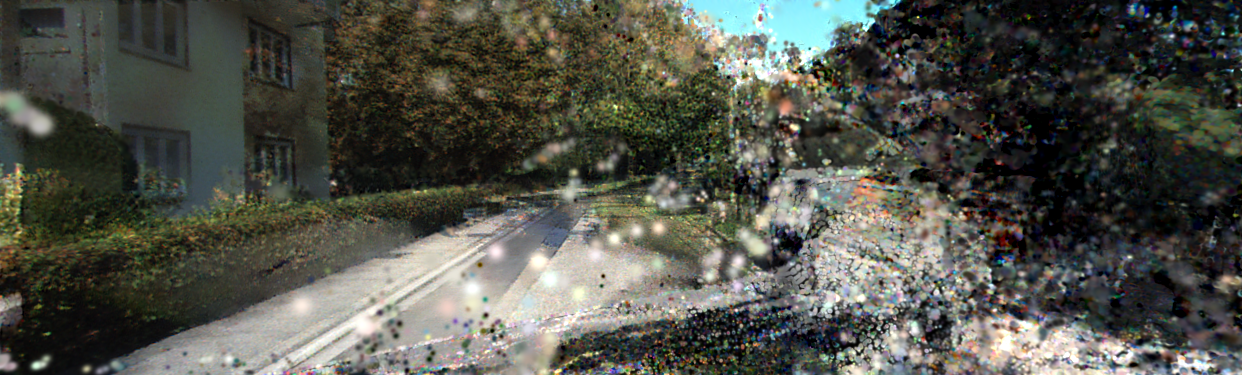} &
\includegraphics[width=0.23\textwidth]{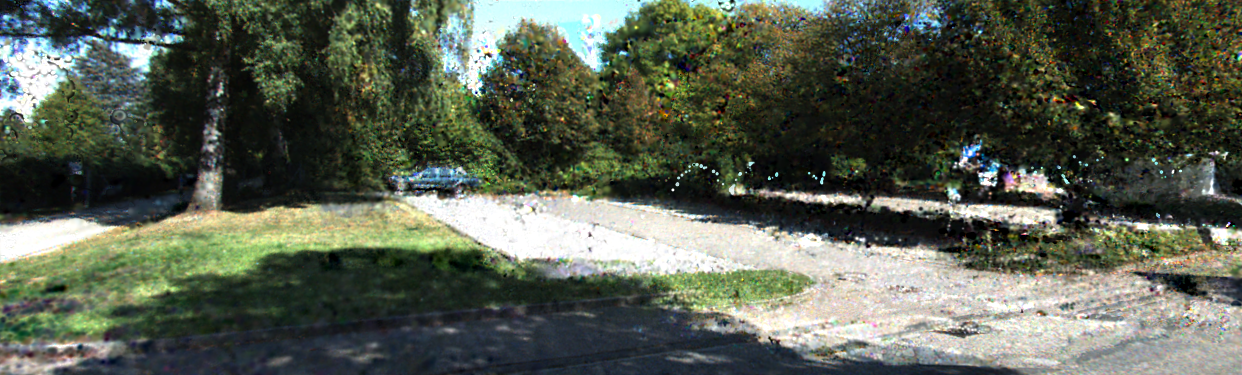} \\
\rotatebox{90}{\hspace{0.3em} \tiny{MonoGS }} &
\includegraphics[width=0.23\textwidth]{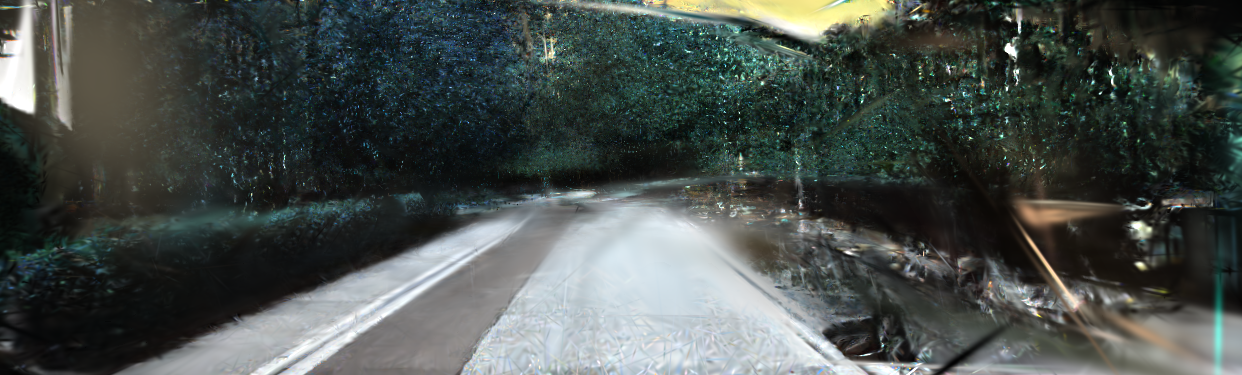} &
\includegraphics[width=0.23\textwidth]{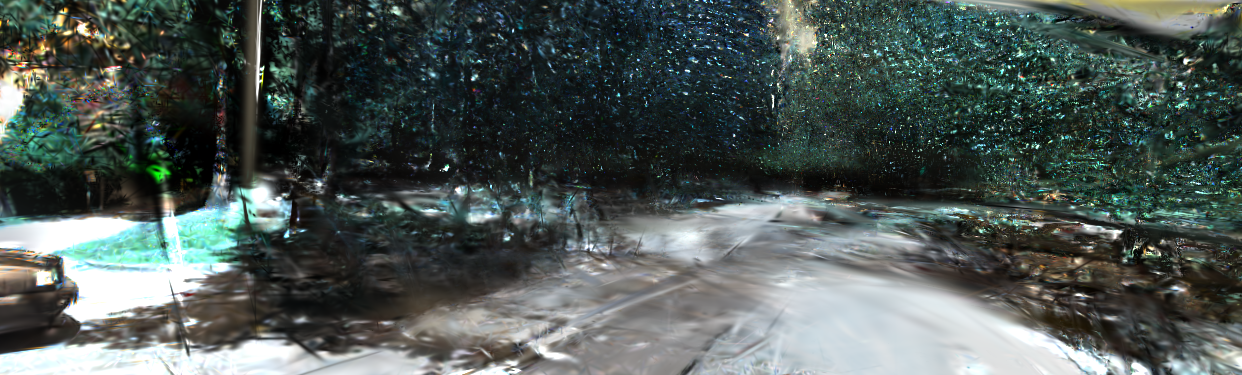} \\
\rotatebox{90}{\tiny{Point-SLAM }} &
\includegraphics[width=0.23\textwidth]{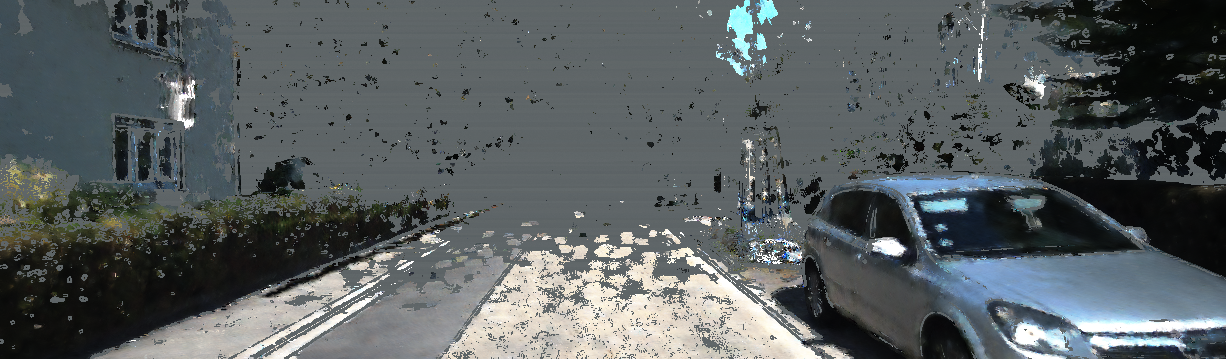} &
\includegraphics[width=0.23\textwidth]{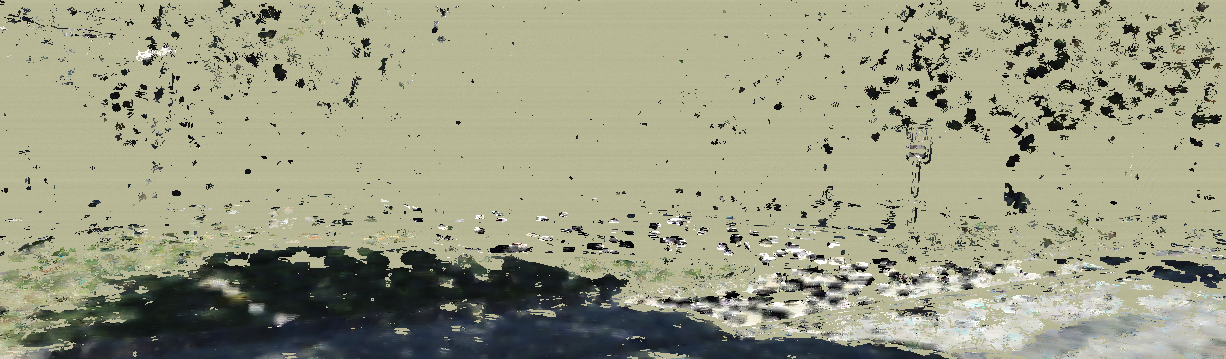} \\
\rotatebox{90}{\tiny{\hspace{0.3em}\textbf{Ours} (LiDAR)}} &
\includegraphics[width=0.23\textwidth]{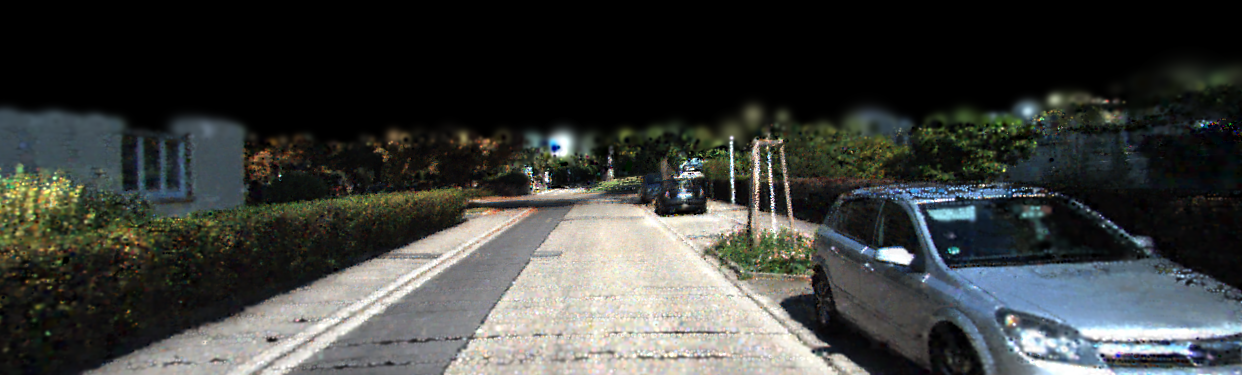} &
\includegraphics[width=0.23\textwidth]{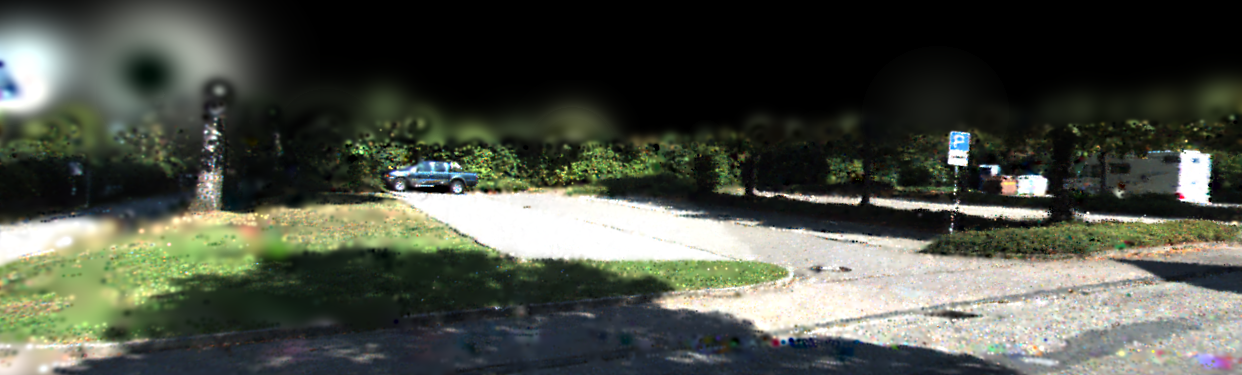} \\
\rotatebox{90}{\hspace{0.8em} \tiny{\textbf{Ours}}} &
\includegraphics[width=0.23\textwidth]{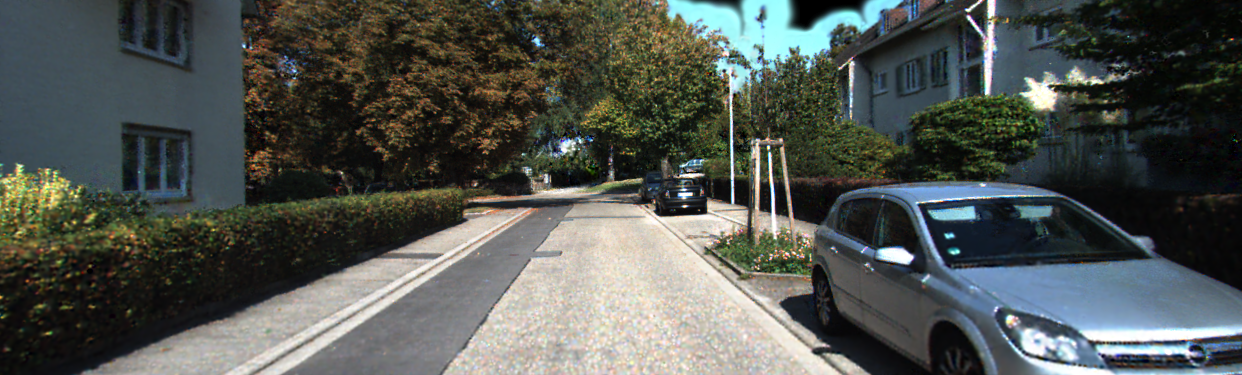} &
\includegraphics[width=0.23\textwidth]{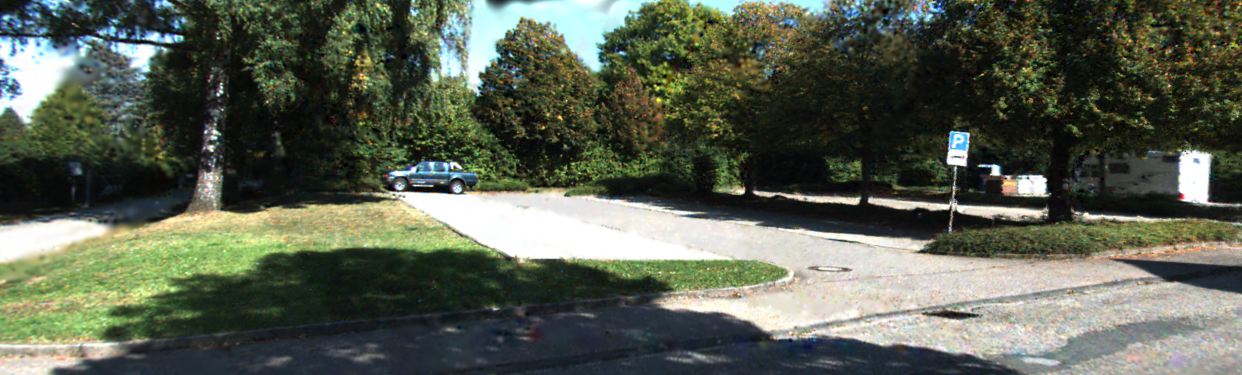} \\
\rotatebox{90}{\hspace{1.1em} \tiny{GT}} &
\includegraphics[width=0.23\textwidth]{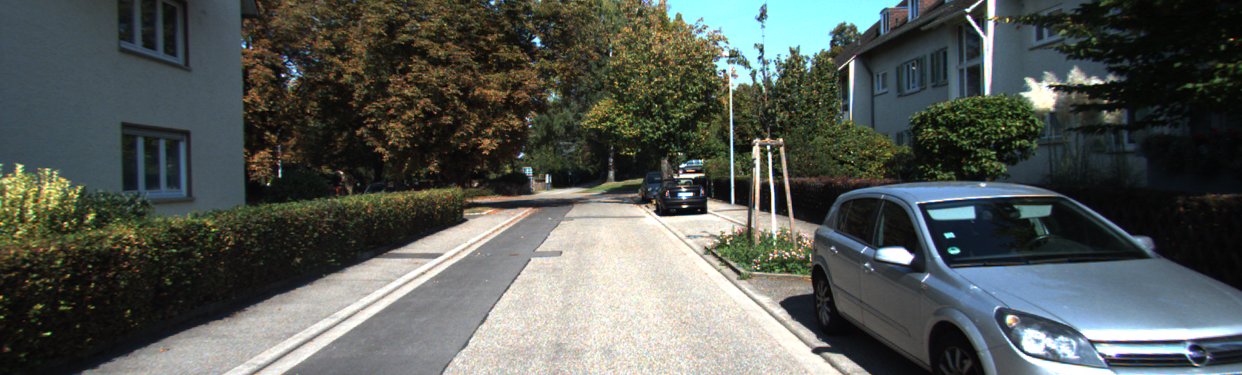} &
\includegraphics[width=0.23\textwidth]{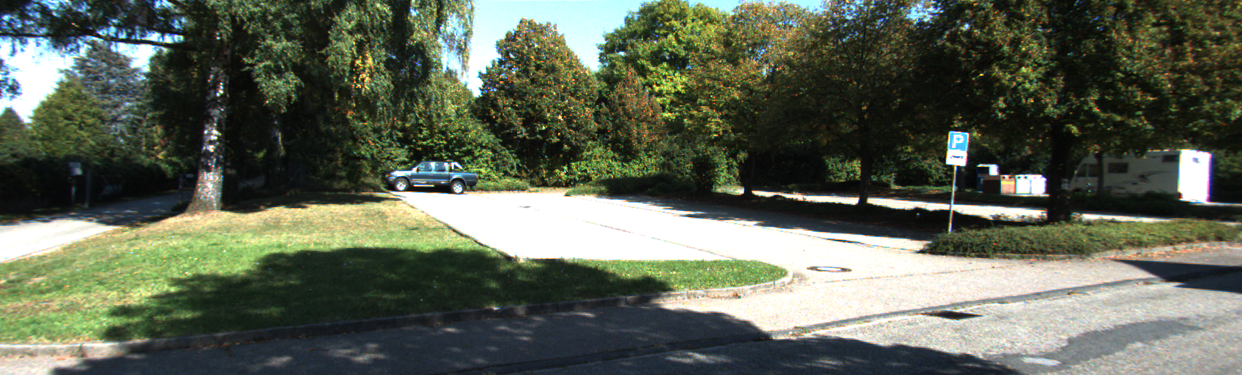} \\
\end{tabular}\vspace{-0.25cm}
\caption{\textbf{Visualization of rendering quality on KITTI.}} 
\label{fig:kitti_illus}
\end{figure}

\begin{table}[t]\small
    \centering
    \setlength{\tabcolsep}{1mm} 
    \begin{tabular*}{1.0\linewidth}{ccccccc}
    \toprule[2pt]
    \specialrule{0em}{1pt}{1pt}
    \multirow{1}*{}{Methods} & {03} & {05} & {06} & {07} & {09} & {10}\\
    \specialrule{0em}{1pt}{1pt}
    \specialrule{0em}{1pt}{1pt}
    \hline
    \specialrule{0em}{1pt}{1pt}
    SplaTAM & \textcolor{red}{\xmark} & \textcolor{red}{\xmark} & \textcolor{red}{\xmark} & \textcolor{red}{\xmark} & \textcolor{red}{\xmark} & \textcolor{red}{\xmark}\\
    \textbf{BGS-SLAM (Ours)} & 8.08 & 45.12 & 16.77 & 14.81 & 20.36 & 20.62 \\
    \bottomrule[2pt]
    \end{tabular*}\vspace{-0.25cm}
    \caption{\textbf{Memory Consumption Analysis (GB).} \textcolor{red}{\xmark} indicates that the method fails to process full sequences, running out of memory after few hundreds frames. }\vspace{-0.25cm}
    \label{Memory}
\end{table}

\subsection{Comparison with State-of-The-Art SLAM}
\textbf{Tracking and Mapping Performance.}  We compare BGS-SLAM against state-of-the-art radiance field-based SLAM methods on the KITTI and KITTI-360 datasets in terms of tracking accuracy and mapping quality. Quantitative results are reported in Table~\ref{results} and Table~\ref{results-360}. Due to memory constraints, methods like SplaTAM~\cite{keetha2024splatam}, MonoGS~\cite{matsuki2024gaussian}, and Point-SLAM~\cite{Sandström2023ICCV} were evaluated only on the first 300 frames per sequence. However, their tracking threads showed large pose estimation errors in outdoor environments, limiting their applicability in real-world large-scale scenes. 
In contrast, our tracking, grounded in classical SLAM pose estimation, provides robust and accurate performance even in complex, large-scale scenarios.

For mapping and view synthesis, BGS-SLAM substantially outperforms all baselines across all visual metrics, achieving an average PSNR improvement of over 6 dB in KITTI dataset. In KITTI-360 dataset, BGS-SLAM achieves more than a 10 dB improvement in PSNR and the best depth reconstruction accuracy with the lowest Depth L1 error.

Fig.~\ref{fig:kitti_illus} illustrates the rendering performance of BGS-SLAM vs. baselines. A LiDAR-supervised variant is also shown, highlighting its advantages over active sensors in outdoor settings. 
Compared to SplaTAM, MonoGS, and Point-SLAM, our method achieves the highest fidelity and continuity in large-scale outdoor scenes. 

\textbf{Memory Efficiency.} Table. \ref{Memory} shows the memory consumption analysis for all methods. SplaTAM fails to process complete KITTI sequences even on an NVIDIA A40 GPU with 48GB of memory, whereas BGS-SLAM succeeds at it. 




\section{Conclusion}
In this paper, we present BGS-SLAM, the first 3DGS-SLAM system for outdoor scenarios using only stereo RGB input. Our novel contributions include leveraging pre-trained deep stereo networks for depth supervision and introducing a multi-loss optimization strategy that combines RGB, depth, normal, and smoothness losses to enhance geometric consistency and novel view synthesis quality. Experiments on KITTI and KITTI-360 demonstrate that BGS-SLAM achieves superior tracking and mapping quality compared to existing radiance-field SLAM approaches without requiring expensive LiDAR sensors.

\textbf{Limitations.} BGS-SLAM does not yet operate in real-time, with average tracking and mapping times of 0.24\,s and 1.37\,s per frame, respectively—posing a limitation for practical SLAM applications. The computational overhead is further increased by the inference time of the deep stereo network, in addition to the iterative optimization of 3D Gaussians for each frame.



\bibliography{aaai2026}

\end{document}